%% file: main.tex
\documentclass[runningheads]{llncs}
\usepackage[T1]{fontenc}
\usepackage{graphicx,verbatim}
\usepackage[hidelinks,colorlinks=true,linkcolor=blue,citecolor=blue]{hyperref}
\usepackage{makecell} 
\usepackage{multirow}
\usepackage{amsmath}
\usepackage[capitalize]{cleveref}
\usepackage{amssymb}
\usepackage{bbding}
\crefname{section}{Sec.}{Secs.}
\Crefname{section}{Section}{Sections}
\Crefname{table}{Table}{Tables}
\crefname{table}{Tab.}{Tabs.}
\usepackage{booktabs}
\usepackage{cite}
\usepackage{ulem}
\graphicspath{{figures/}}

\begin{document}
\title{Echo-Path: Pathology-Conditioned Echo Video Generation}
\author{
Kabir Hamzah Muhammad\inst{1} \and
Marawan Elbatel\inst{1} \and
Yi Qin\inst{1} \and
Xiaomeng Li\inst{1}
}

\authorrunning{K. H. Muhammad et al.} 

\institute{
    Hong Kong University of Science and Technology, Hong Kong \\
    \email{\{khmuhammad, mkfmelbatel, yqinar, eexmli\}@connect.ust.hk}
}

\maketitle             
\begin{abstract}
Cardiovascular diseases (CVDs) remain the leading cause of mortality globally, and echocardiography is critical for diagnosis of both common and congenital cardiac conditions. However, echocardiographic data for certain pathologies are scarce, hindering the development of robust automated diagnosis models. In this work, we propose Echo-Path, a novel generative framework to produce echocardiogram videos conditioned on specific cardiac pathologies. Echo-Path can synthesize realistic ultrasound video sequences that exhibit targeted abnormalities, focusing here on atrial septal defect (ASD) and pulmonary arterial hypertension (PAH). Our approach introduces a pathology-conditioning mechanism into a state-of-the-art echo video generator, allowing the model to learn and control disease-specific structural and motion patterns in the heart. Quantitative evaluation demonstrates that the synthetic videos achieve low distribution distances, indicating high visual fidelity. Clinically, the generated echoes exhibit plausible pathology markers. Furthermore, classifiers trained on our synthetic data generalize well to real data and, when used to augment real training sets, it improves downstream diagnosis of ASD and PAH by 7\% and 8\% respectively. Code, weights and dataset are available \href{https://github.com/Marshall-mk/EchoPathv1/tree/main}{here.}

\end{abstract}

\input{Section_1_Introduction}

\input{Section_2_Methodology}

\input{Section_3_Experiments}
\input{Section_4_Conclusion}

\bibliographystyle{splncs04}
\bibliography{references}

\end{document}

%% file: Section_1_Introduction.tex
\section{Introduction}
Cardiovascular diseases (CVDs) are the leading global cause of mortality, responsible for over 18 million deaths annually~\cite{cite1}. Echocardiography (echo) is a key diagnostic tool for assessing cardiac anatomy and function due to its portability, real-time imaging, and non-invasive nature. It plays a vital role in detecting congenital defects such as atrial septal defect (ASD), and chronic conditions like pulmonary arterial hypertension (PAH), both of which affect right heart structure and interventricular septal motion~\cite{cite2, cite3}. However, acquiring diverse and well-labeled echocardiographic data—especially for rare pathologies—is challenging due to both clinical scarcity and patient privacy constraints.

To address data scarcity and privacy issues in medical imaging, generative models have gained significant attention. Traditional generative adversarial networks (GANs) have been applied to ultrasound synthesis~\cite{cite5}, but often produce low-fidelity results. More recently, diffusion models have emerged as a powerful alternative, offering improved sample quality and training stability. In the context of echocardiography, the EchoNet-Synthetic pipeline~\cite{cite4} introduced a latent video diffusion approach for generating realistic apical four-chamber (A4C) cine loops conditioned on ejection fraction (EF). However, it does not explicitly model discrete structural abnormalities (e.g., septal defects).

Other generative approaches, such as ECHOPulse~\cite{cite10} and HeartBeat~\cite{cite9}, have introduced additional forms of conditioning—e.g., ECG signals, segmentation maps, or textual prompts—to guide generation. While these methods enable greater control, they often require additional inputs or annotations, limiting their applicability to broad pathology-specific synthesis. Similarly, Free-Echo~\cite{cite6} proposes test-time conditional sampling using noisy segmentation masks, but it does not learn an explicit disease-aware latent space.

In this work, we introduce \textbf{Echo-Path}, a novel generative framework specifically designed for pathology-conditioned echocardiography video synthesis. Echo-Path explicitly integrates Echo-Specific attributes into the generative modeling process, enabling targeted control over anatomical appearance and cardiac motion dynamics. Our contributions are as follows:

\begin{itemize}
\item We present a diffusion-based framework conditioned on pathology and anatomy, enabling precise control of pathological features in generated videos.

\item We introduce an autoregressive sampling method that maintains pathology-specific motion across extended echo video sequences.

\item Echo-Path generates pathology-specific videos with clinically meaningful features, validated through both quantitative and qualitative metrics.
\end{itemize}

%% file: Section_2_Methodology.tex
\section{Methods}
\subsection{Echo-Path Generative Framework}
\begin{figure*}[t]
    \centering
    \includegraphics[width=\textwidth]{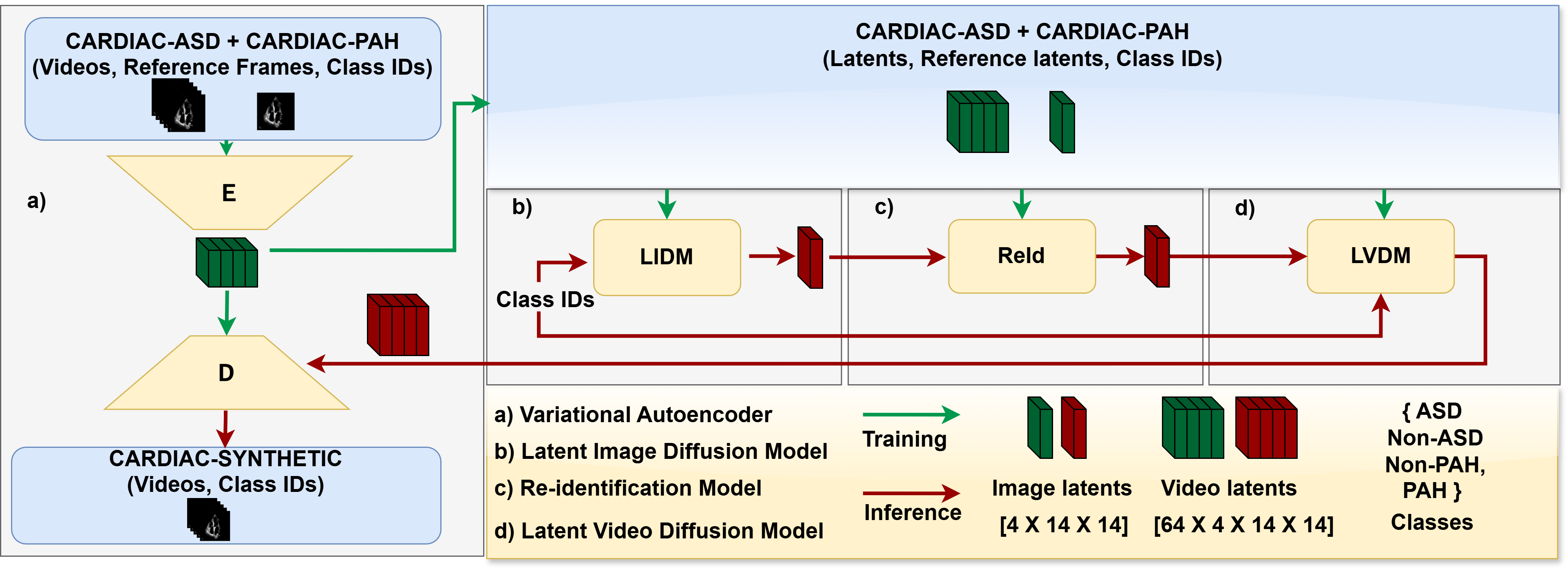}
    \caption{Schematic of the Echo-Path class-conditioned video generation framework. Green arrows denote training flows and red arrows indicate the inference pipeline.}
    \label{fig:framework}
\end{figure*}

\noindent\textbf{Overview:} Echo-Path is a multi-stage, class-conditioned diffusion pipeline that synthesizes realistic echo videos reflecting specific pathologies. As shown in Fig.~\ref{fig:framework}, first, a Latent Image Diffusion Model (LIDM) generates a single representative cardiac frame in a low-dimensional latent space. Then a re-identification model checks the generated frame for privacy. A Latent Video Diffusion Model (LVDM) produces a sequence of subsequent frames given the privacy-compliant frame, capturing pathology-specific motion dynamics. Finally, a Variational Autoencoder (VAE) decodes the generated latent frames back to image space. By explicitly conditioning on disease class labels, our method can generate diagnostically meaningful differences in both structure and motion. We build on the EchoNet-Synthetic framework \cite{cite4} by reusing its variational autoencoder and privacy filtering components, but we extend it with discrete pathology conditioning and an autoregressive video generation strategy to enable coherent long-form sequences.

\noindent\textbf{Latent Image Diffusion (LIDM):} We first project all echo frames into a $d$-dimensional latent space using the pre-trained VAE (from \cite{cite4}) for efficiency. The LIDM is a 2D diffusion model defined in this latent image space. Starting from Gaussian noise $z_T \sim \mathcal{N}(0,I)$, the LIDM learns to denoise toward a realistic initial frame $z_{0}$ representative of a given condition. Conditioning is provided by the pathology label (one of ASD, Non-ASD, PAH or Non-PAH), which is fed into the diffusion UNet via cross-attention. This diffusion process has a closed-form:
\[
x_t = \sqrt{\bar{\alpha}_t}\,x_0 + \sqrt{1 - \bar{\alpha}_t}\,\epsilon,
\]
where $\bar{\alpha}_t = \prod_{i=1}^t (1-\beta_i)$ and $\epsilon \sim \mathcal{N}(0,I)$. The reverse model is trained to predict and remove the added noise, reconstructing $x_0$ from $x_T$. We adopt the $v$-prediction parameterization~\cite{cite16} for training the LIDM, which improves convergence and sample fidelity. During training we occasionally drop the class conditioning (classifier-free guidance augmentation) so that at inference we can apply classifier-free guidance (CFG) to steer generation towards class-specific features. Using an appropriate guidance scale, the LIDM sampler produces a latent frame $z_{\text{heart}}$ that depicts plausible anatomy for the target diagnosis. We also apply a re-identification based filter (as in \cite{cite4}) to ensure $z_{\text{heart}}$ is not too similar to any real training frame, preserving patient privacy.

\noindent\textbf{Latent Video Diffusion (LVDM):} Given the conditioned latent image $z_{\text{heart}}$, the LVDM generates a 64-frame video in latent space that begins with this initial frame and evolves with realistic cardiac motion. The LVDM is a diffusion model over spatio-temporal latent tensors $Z \in \mathbb{R}^{d\times 64}$. To preserve the initial anatomy throughout generation, we supply $z_{\text{heart}}$ as an extra input to the model for every frame in the sequence (i.e., by concatenating the latent $z_{\text{heart}}$ to each frame’s latent representation along the channel dimension). The disease label is also provided as a global condition. Starting from spatio-temporal Gaussian noise $Z_T$, the LVDM performs $T$ denoising steps to produce $Z_{0:63}$, a coherent 64-frame latent sequence. Crucially, the first frame of $Z_{0:63}$ aligns with $z_{\text{heart}}$ (ensuring structural consistency), and the subsequent frames exhibit pathology-specific dynamics driven by the class condition (for example, a PAH-conditioned sequence may show interventricular septum flattening and right ventricular dilation over the cardiac cycle). We apply CFG during this video generation as well, to emphasize subtle disease-related motion cues. The result after diffusion is a 64-frame latent video $Z_{0:63}$ capturing one or more heartbeats consistent with the specified condition.

\subsection{Autoregressive Sampling} 
Generating long, temporally coherent echo  videos is challenging due to the high-dimensional spatio-temporal complexity of cardiac motion and the fixed-length constraints of diffusion models. Many prior methods such as\cite{cite40}\cite{cite5}\cite{cite6}\cite{cite7}\cite{cite10} generate only fixed-length sequences, limiting their practical use for longer video synthesis. Others, like \cite{cite4}, extend videos by stitching separately generated segments, but this approach is computationally expensive at inference. To overcome these issues, we propose a blockwise autoregressive strategy that sequentially generates video segments conditioned on the previous segments. This ensures smooth temporal transitions and preserves pathology-specific motion dynamics over extended durations while reducing inference complexity. Specifically,
let \(Z^{(k)}\) denote the \(k\)-th 64-frame latent video block generated by the LVDM. The initial block \(Z^{(0)}\) is generated conditioned on the latent image \(z_{\text{heart}}\) from the LIDM and class label \(y\):

\[
Z^{(0)} \sim p_\theta(Z \mid z_{\text{heart}}, y).
\]

Each subsequent block \(Z^{(k)}\) (for \(k \geq 1\)) is generated conditioned on the last frame of the previous block \(z^{(k-1)}_{63}\) and the same class label:

\[
Z^{(k)} \sim p_\theta(Z \mid z^{(k-1)}_{63}, y).
\]

This ensures temporal continuity, as the first frame of each new block equals the last frame of the previous block (\(z^{(k)}_0 = z^{(k-1)}_{63}\)). The full video \(\hat{Z}\) is formed by concatenating these blocks:

\[
\hat{Z} = \bigcup_{k=0}^{M-1} Z^{(k)}.
\]

%% file: Section_3_Experiments.tex
\section{Experiments}
\noindent\textbf{Dataset:}
We utilize the Cardiac-ASD and Cardiac-PAH echocardiography datasets introduced by \cite{cite14}. The Cardiac-ASD set contains 207 apical-4-chamber videos (ASD vs. Non-ASD labels), and Cardiac-PAH includes 448 videos (PAH vs. Non-PAH labels). We used the prescribed training/validation/test splits as in \cite{cite14}. Each volumetric study is converted to a 2D frame sequence and resampled to 32~fps. All frames are resized to $112\times112$ and duplicated across three channels to create pseudo-RGB inputs, matching the format of prior work \cite{cite4}.

\noindent\textbf{Implementation Details:} 
The VAE was reused directly from the pre-trained EchoNet-Synthetic model \cite{cite4} to leverage its established latent encoding capabilities. The LIDM was trained for 5,000 steps using a batch size of 1,024 distributed across 4 NVIDIA RTX 3090 GPUs. We used the Adam optimizer with a learning rate of \(3 \times 10^{-4}\). The re-identification model was trained on a single RTX 3090 GPU with a batch size of 128 for 1,000 epochs to ensure effective identity filtering during generation. The LVDM was trained for 100,000 iterations with a batch size of 16 distributed across 4 RTX 3090 GPUs. A learning rate of \(1 \times 10^{-4}\) was used, with the Adam optimizer applied throughout. 

\subsection{Performance Evaluation}
\noindent\textbf{Baselines:} We first establish baseline performance on the real echocardiography data. The current state of the art (SOTA) model for ASD and PAH detection is the CardiacNet model from~\cite{cite14}. It is a specialized network designed for ASD and PAH detection, and achieves strong accuracy of approximately 89.6\% on ASD and 85.7\% on PAH. However, CardiacNet is computationally intensive, requiring training over 1,000 epochs on multiple GPUs and several days of continuous training, which limits its practicality for rapid experimentation.

To provide a more accessible reference point, we trained a simpler ResNet-based video classifier on the same real training set. Despite its reduced complexity, this model attained respectable accuracies of 84.8\% for ASD and 78.3\% for PAH on the real test set. We also established baselines using a recent study~\cite{cite41} that introduced a complete AI-enabled echocardiography interpretation with multi-task deep learning, here we utilized it's binary classification head. These results serve as pragmatic benchmarks for evaluating the effectiveness of models trained on synthetic data, balancing performance with computational efficiency.

\noindent\textbf{Synthetic Video Generation Quality:} We assess the realism of generated echo videos using distribution similarity metrics. In particular, we compute FID, FVD, and IS~\cite{cite22,cite23,cite24}. Table~\ref{tab:lvdm_results} compares our Echo-Path generative framework to the EchoNet-Synthetic (``Echosyn'') baseline~\cite{cite4} under two inference settings: low guidance ($w=1$) versus high guidance ($w=5$). Lower FID/FVD and higher IS indicate better fidelity/diversity. Our method achieves consistently lower FID than Echosyn (e.g., 18.35 vs 20.05 with strong guidance), suggesting our synthetic samples are closer to the real data distribution. The \textbf{FVD\(_{16}\)} results reveal moderate yet consistent gains for Echo-Path.  
With strong guidance ($w\!=\!5$) our model lowers FVD by $0.8\%$ on ASD (127.49 vs.\ 128.28) and by $7.6\%$ on PAH (143.42 vs.\ 155.26).  Under weak guidance ($w\!=\!1$) the scores are essentially tied on ASD (135.24 vs.\ 134.88) while still $8.3\%$ better on PAH (144.58 vs.\ 156.95).  
Lower FVD values indicate that Echo-Path generates sequences with equal or superior temporal coherence relative to the Echosyn baseline. 
The IS values are around 2.0 for all cases, reflecting the limited diversity of the dataset. Overall, Echo-Path delivers equal or better realism compared to the prior state-of-the-art.

\begin{table}[t]
  \centering
  \renewcommand{\tabcolsep}{5pt}
  \caption{Comparison of generative methods and guidance scales ($w$) on the ASD/PAH data. Bold indicates best performance, underlined indicates second best.}
  \label{tab:lvdm_results}
  \begin{tabular}{lcccccc}
    \toprule
    & \multicolumn{3}{c}{ASD} & \multicolumn{3}{c}{PAH} \\ 
    \cmidrule(lr){2-4} \cmidrule(lr){5-7}
                            & FID ↓ & FVD$_{16}$ ↓ & IS ↑ & FID ↓ & FVD$_{16}$ ↓ & IS ↑ \\
    \midrule
    Echosyn~\cite{cite4}, $w=1$ & 20.55 & 134.88 & $2.06\!\pm\!0.08$ & 24.36 & 156.95 & \uline{$2.02\!\pm\!0.09$} \\
    Echosyn~\cite{cite4}, $w=5$ & 20.05 & \uline{128.28} & $2.06\!\pm\!0.11$ & 24.25 & 155.26 & $1.99\!\pm\!0.09$ \\
    Ours, $w=1$                 & \uline{18.87} & 135.24 & \uline{$2.06\!\pm\!0.08$} & \uline{22.42} & \uline{144.58} & $2.01\!\pm\!0.07$ \\
    Ours, $w=5$                 & \textbf{18.35} & \textbf{127.49} & $\mathbf{2.06\!\pm\!0.07}$ & \textbf{22.23} & \textbf{143.42} & $\mathbf{2.05\!\pm\!0.10}$ \\
    \bottomrule
  \end{tabular}
\end{table}

\begin{figure*}[t]
    \centering
    \includegraphics[width=\textwidth, height=5cm]{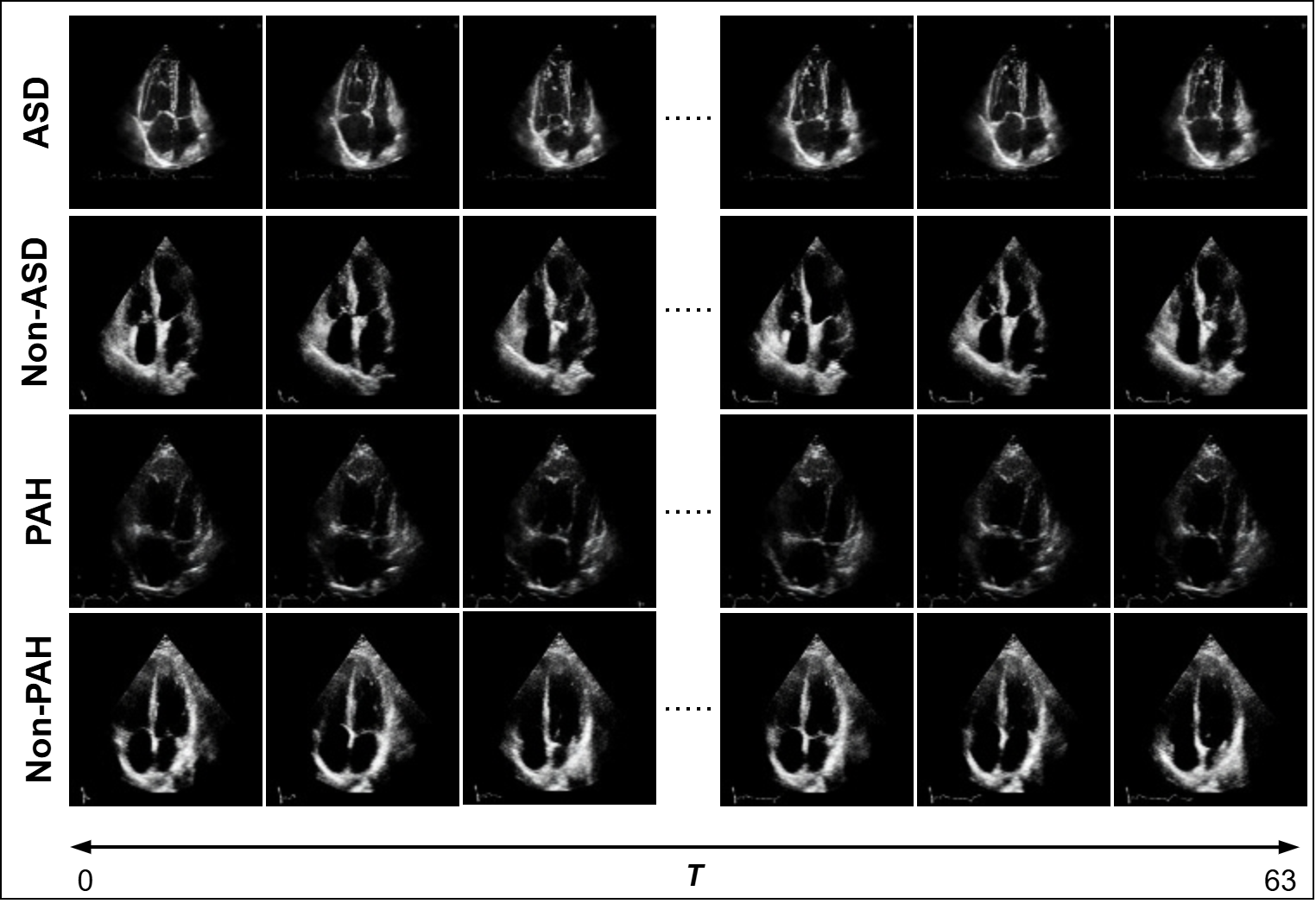}
    \caption{Example frames from class-conditioned synthetic echo videos (ASD and PAH).}
    \label{fig:synthetic_video_frames}
\end{figure*}

Qualitatively, the generated videos appear realistic and capture the intended pathology traits. Figure~\ref{fig:synthetic_video_frames} shows frames from synthetic ASD and PAH examples. The ASD example exhibits an enlarged right atrium and a clear septal defect, while the PAH example shows a bowed interventricular septum and dilated right ventricle. The motion in both videos is smooth and natural, with no obvious artifacts, indicating that the diffusion models successfully learned to produce plausible cardiac motion conditioned on the disease.

\begin{figure*}[t]
    \centering
    \includegraphics[width=\textwidth, height=5cm]{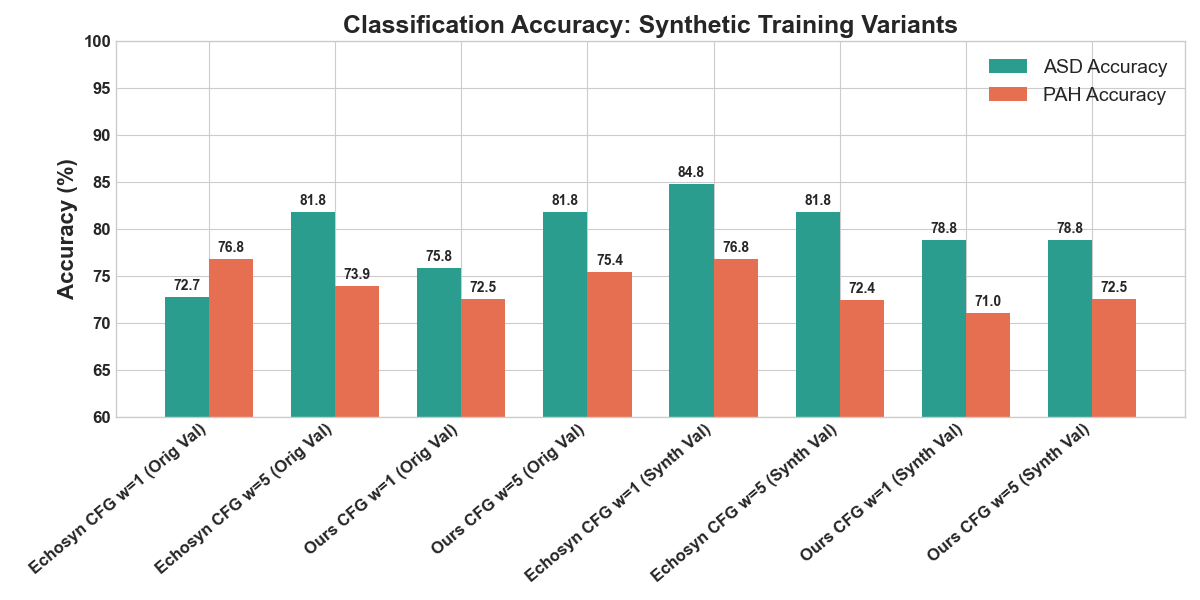}
    \caption{Classification results on real test data when training on synthetic videos with real validation set (orig val) and with synthetic validation set (synth val). We compare classifiers trained on EchoNet-Synthetic vs.\ Echo-Path generated data.}
    \label{fig:variants}
\end{figure*}

\noindent\textbf{Cross-Domain Classification Performance:} We next evaluate how well classifiers trained on synthetic data perform on real echocardiograms. We train the ResNet classifier using only generated videos as training data and test on the real hold-out test set. We consider two scenarios: (1) using a small set of real validation videos during training (to guide model selection), and (2) using only synthetic data for both training and validation (a zero-real-data scenario). Table~\ref{tab:cls_results_st_ov_ot} summarizes the results for the first scenario (synthetic train with real val), and we observed similar trends in the fully synthetic scenario (results in Fig~\ref{fig:variants}). For ASD classification, a model trained on our Echo-Path synthetic videos achieved up to 81.8\% accuracy on real test data when using guided generation ($w=5$). This matches the best accuracy obtained using the baseline Echosyn data (also 81.8\% with $w=5$). Without guidance ($w=1$), accuracies drop to 75.8\% (Ours) vs 72.7\% (Echosyn). For PAH, synthetic-trained models reached slightly lower performance: the best test accuracy was 76.8\%, achieved by the baseline synthetic data without guidance, while our method attained 75.4\% with guidance. Generally, using a higher guidance weight during generation improved the ASD synthetic training results substantially (e.g., +6–9\% accuracy over $w=1$), whereas for PAH the effect was smaller or even negative for the baseline. The gap between synthetic-trained and real-trained classifiers is more pronounced in PAH, suggesting that the synthetic PAH videos may not capture the full complexity of real severe PAH cases. Nonetheless, the fact that an ASD classifier trained purely on synthetic data can reach $>$80\% of real-world accuracy is a strong indication that our generative model has learned the key features of ASD.

\begin{table}[t]
\centering
\renewcommand{\tabcolsep}{5pt}
\caption{Classification results on real test data when training on synthetic videos (with a small set of real validation data). We compare classifiers trained on EchoNet-Synthetic vs.\ Echo-Path generated data. Best values in bold, second best underlined.}
\label{tab:cls_results_st_ov_ot}
\begin{tabular}{lcccccc}
\toprule
& \multicolumn{3}{c}{ASD} & \multicolumn{3}{c}{PAH} \\ 
\cmidrule(lr){2-4} \cmidrule(lr){5-7}
 & ACC $\uparrow$ & F1 $\uparrow$ & AUC-ROC $\uparrow$ & ACC $\uparrow$ & F1 $\uparrow$ & AUC-ROC $\uparrow$ \\ 
\midrule
Echosyn\cite{cite4}, $w=1$ & 0.727 & 0.718 & 0.722 & \textbf{0.768} & 0.675 & 0.659 \\
Echosyn\cite{cite4}, $w=5$ & \uline{0.818} & \uline{0.813} & \uline{0.814} & 0.739 & \textbf{0.683} & \textbf{0.683} \\
Ours, $w=1$ & 0.758 & 0.746 & 0.752 & 0.725 & 0.660 & 0.658 \\
Ours, $w=5$ & \textbf{0.818} & \textbf{0.816} & \textbf{0.816} & \uline{0.754} & \uline{0.675} & \uline{0.664} \\
\bottomrule
\end{tabular}
\end{table}

In the completely zero-shot setting (no real validation data), we found a similar pattern: the synthetic-trained ASD classifier achieved up to 78.8\% accuracy (ours) vs 84.8\% (baseline) on real test, while PAH accuracy remained in a comparable range (72–76\% for both methods). These results show that while a few real validation samples help tune the models, performance is mainly limited by the domain gap in synthetic data—especially for PAH. Though realistic, synthetic PAH videos may miss subtle variations in real hearts, leading to weaker generalization.

\noindent\textbf{Augmented Training with Synthetic Data.} Finally, we evaluate the benefit of using synthetic videos to augment the real training set. We combine the Echo-Path generated videos with the original real training data and train the ResNet classifier on this augmented dataset. Table~\ref{tab:cls_results_aug} compares the augmented training results with the original benchmarks. Incorporating synthetic examples yields clear improvements. For ASD, the augmented model achieves 91.8\% test accuracy with an AUC-ROC of 0.905, surpassing both the real-data-only ResNet baseline (84.8\% ACC) and even slightly exceeding the complex CardiacNet model’s performance (89.6\% ACC, 0.912 AUC in~\cite{cite14}). For PAH, training with augmented data raises the test accuracy to 86.3\% (vs 78.3\% with real-only). The PAH AUC-ROC with augmentation is 0.830, which is below the CardiacNet’s 0.893, indicating there is still room to improve in capturing all PAH-related cues. Overall, using our synthetic videos as additional training data significantly boosts the classifier’s metrics for both tasks, demonstrating the practical value of Echo-Path. Notably, these gains are achieved with a relatively simple ResNet, highlighting how augmenting data can outperform more complex models trained on limited real data.

\begin{table}[t]
\centering
\renewcommand{\tabcolsep}{1pt}
\caption{Classification results on real test data when training with augmented (real + synthetic) datasets. We compare our augmented ResNet model with the original CardiacNet benchmark~\cite{cite14}, Pan-Echo~\cite{cite41}, and the baseline ResNet model.}
\label{tab:cls_results_aug}
\begin{tabular}{lcccc}
\toprule
 & \multicolumn{2}{c}{ASD} & \multicolumn{2}{c}{PAH} \\ 
\cmidrule(lr){2-3} \cmidrule(lr){4-5}
 & ACC $\uparrow$ & AUC-ROC $\uparrow$ & ACC $\uparrow$ & AUC-ROC $\uparrow$ \\
\midrule
Baseline (real only) & 0.848 & 0.846 & 0.783 & 0.729 \\
Pan-Echo\cite{cite41} (real only) & 0.734 & 0.734 & 0.766 & 0.761 \\
CardiacNet\cite{cite14} (real only) & \uline{0.896} & \textbf{0.912} & \uline{0.857} & \textbf{0.893} \\
Ours (augmented) & \textbf{0.918} & \uline{0.905} & \textbf{0.863} & \uline{0.830} \\
\bottomrule
\end{tabular}
\end{table}

%% file: Section_4_Conclusion.tex
\section{Conclusion}
We introduced Echo-Path, a pathology-conditioned video generation framework for realistic cardiac ultrasound videos. By combining latent image and video diffusion models with a privacy-preserving filter, Echo-Path generates synthetic echoes resembling real patient data while offering pathology control. Quantitatively, Echo-Path achieved low distribution distances (FID $\approx$18) and trained classifiers approaching real-data performance. When used to augment limited real datasets, it significantly improved ASD and PAH classification accuracy (91.8\% and 86.3\%, respectively). These results demonstrate that diffusion-based generative models can address data scarcity in echocardiography and enhance training for rare conditions, without compromising privacy.

\textbf{Prospect of application:} Echo-Path offers significant potential for improving echocardiographic assessments, especially for rare cardiac conditions. By augmenting datasets with synthetic videos, it can address data scarcity, improve diagnostic model training, and support robust AI applications in clinical settings, enhancing disease detection and patient care.